\documentclass{article} 
\usepackage{ICLR2020,times}

\usepackage{amsmath,amsfonts,bm}

\def\eqref#1{equation~\ref{#1}}

\def\1{\bm{1}}

\DeclareMathAlphabet{\mathsfit}{\encodingdefault}{\sfdefault}{m}{sl}
\SetMathAlphabet{\mathsfit}{bold}{\encodingdefault}{\sfdefault}{bx}{n}

\usepackage{amsmath,amssymb,amsfonts}
\usepackage{algorithmic}
\usepackage{graphicx}
\usepackage{textcomp}
\usepackage{xcolor}
\usepackage{float}
\usepackage{multirow}
\usepackage{mwe}
\usepackage{verbatimbox}

\usepackage{hyperref}
\usepackage{url}

\title{Early disease diagnosis for Rice Crop}

\author{M. Hammad Masood\thanks{Corresponding Author} , Habiba Saim, Murtaza Taj and  Mian M. Awais  \\
Department of Computer Science, LUMS University, Lahore, Pakistan \\
\texttt{\{17030041, habiba.saim, murtaza.taj, awais\}@lums.edu.pk}
}

\iclrfinalcopy 
\def\BibTeX{{\rm B\kern-.05em{\sc i\kern-.025em b}\kern-.08em
    T\kern-.1667em\lower.7ex\hbox{E}\kern-.125emX}}

\begin{document}

\maketitle

\begin{abstract}

Many existing techniques provide automatic estimation of crop damage due to various diseases.  However, early detection can prevent or reduce the extend of damage itself. The limited performance of existing techniques in early detection is lack of localized information. We instead propose a dataset with annotations for each diseased segment in each image. Unlike existing approaches, instead of classifying images into either healthy or diseased, we propose to provide localized classification for each segment of an images. Our method is based on Mask R-CNN and provides location as well as extend of infected regions on the plant. Thus the extend of damage on the crop can be estimated. Our method has obtained overall $87.6\%$ accuracy on the proposed dataset as compared to  $58.4\%$ obtained without incorporating localized information.

\end{abstract}



\section{Introduction and Background}
Arable land and water are considered as principal natural resources in many developing and underdeveloped countries. In many case it contributes about $18.9\%$ percent of Gross Domestic Product (GDP) and absorbs  $42.3\%$  of  employed  labor  force \citep{PBS2019} and is an important source of foreign exchange earnings.  It feeds whole rural and urban population. For e.g. Pakistan is in the list of top rice exporter in the world. However, agriculture\textquotesingle s contribution in country\textquotesingle ’s gross GDP has declined from $54\%$ to $24.6\%$ over the last decade, and total crop production is almost $50\%$ below its  potential \citep{FinanceSurvey2018}. The  agriculture  sector is facing these issues due to inadequate  agriculture research centers, lack of use of modern agriculture technology and insufficient state of the art facilities. Major reasons of crops yield decline are farmer\textquotesingle’s poor financial position, soil erosion, improper soil restoration, water wastage, old cultivation and harvesting methods,  water logging and salinity, lack of irrigation facilities, defective land reforms, communication gap between farmers and  experts,  plant  diseases, natural  calamities,  and  inefficient and late plant disease detection. Hazardous effects of crop diseases are yield reduction, low quality product, increase in production cost, and ultimately decreased output. Manual monitoring method using naked eyes observation involves large team of experts and continuous monitoring of plants, is a laborious task with low accuracy. High dependency of our economic growth on rice, pushes us to make efforts to protect rice from further damage with the help of an early disease detection system.


 Many researchers have contributed for ICT intervention in the development of such solutions that can help farmers for crop protection and yield improvement \citep{Joshi2016, Phadikar2013}. Some have applied machine learning (ML) and deep learning (DL) models for disease identification. Machine learning techniques applied for classification and identification of crop diseases include Support Vector Machines using spectral vegetation indices \citep{Rumpf2010}, radial kernel function \citep{Hallu2017} for Sugar beet diseases identification and Minimum Distance Classifier (MDC) and K-NN classification methods for rice plant images~\citep{Joshi2016}. Similarly, Important feature selection (reduct - reduced feature set) through Rough Set Theory (RST) has also been used for disease classification of infected rice plants \citep{Phadikar2013}.

 Despite the recent progress in DL, classification/identification of rice diseases using real time field images still possess several challenges. Data collection, pre-processing, handling unfavorable light conditions and occlusion are some prominent challenges that the researchers have to face \citep{b1}. DL is the fastest growing field in ML and provides solutions to variety of computer vision and natural language processing problems. A survey of DL application in agriculture \citep{DL2018} claims that being a modern technique with providing promising results in image processing and data analysis. DL based plant disease framework is used to detect and identify diseases and pests presence in tomato plants \citep{A2017} and for $14$ species of crop, to diagnose twenty six different diseases \citep{Zhongzhi2019}. Convolutional Neural Network (CNN) is the most popular DL architecture for image classification problems. Some researchers have used CNN in combination with other techniques for more complex problems like CNN+HistNN using RGB histograms for crop type identification using UAV imagery. CNNs has also been used to detect maize plant disease named northern leaf blight \citep{Dechant2017}.

Most of the existing literature on crop disease identification particularly for rice is based on vanilla image classification~\citep{Joshi2016, Phadikar2013}. This requires images with significant content of diseased part of rice crop and thus only works at a later stage of infection. We instead propose an early detection technique by simultaneously solving localization and classification. Our approach is based on Mask-R-CNN~\citep{MaskRCNN2017} which segment diseased part of the plant and provides a fine-grained information about crop health (see Sec.~\ref{sec:proposed_method}). We also propose a novel dataset for rice disease classification with detailed annotations for each infected part in the image (see Sec.~\ref{sec:data_collection}). 

\section{Field Data Collection}\label{sec:data_collection}

\begin{figure}[htbp]
	\begin{center}
	\begin{tabular}{cc}
  		\includegraphics[width=.5\columnwidth]{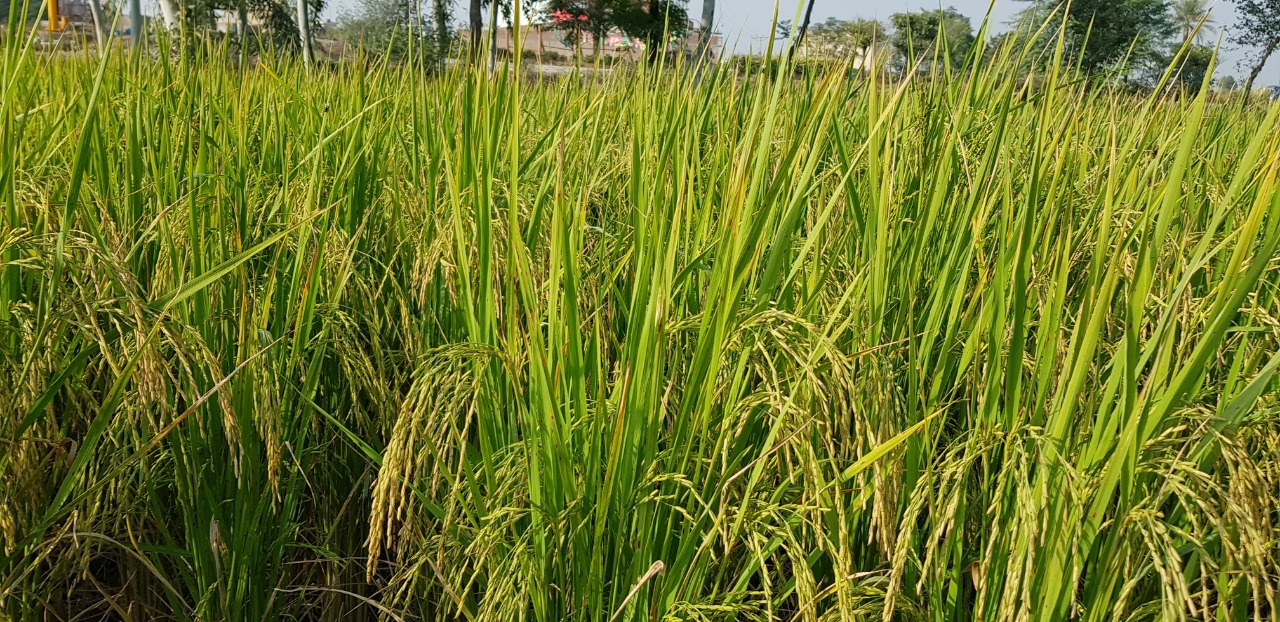} &
  		\includegraphics[width=.41\columnwidth]{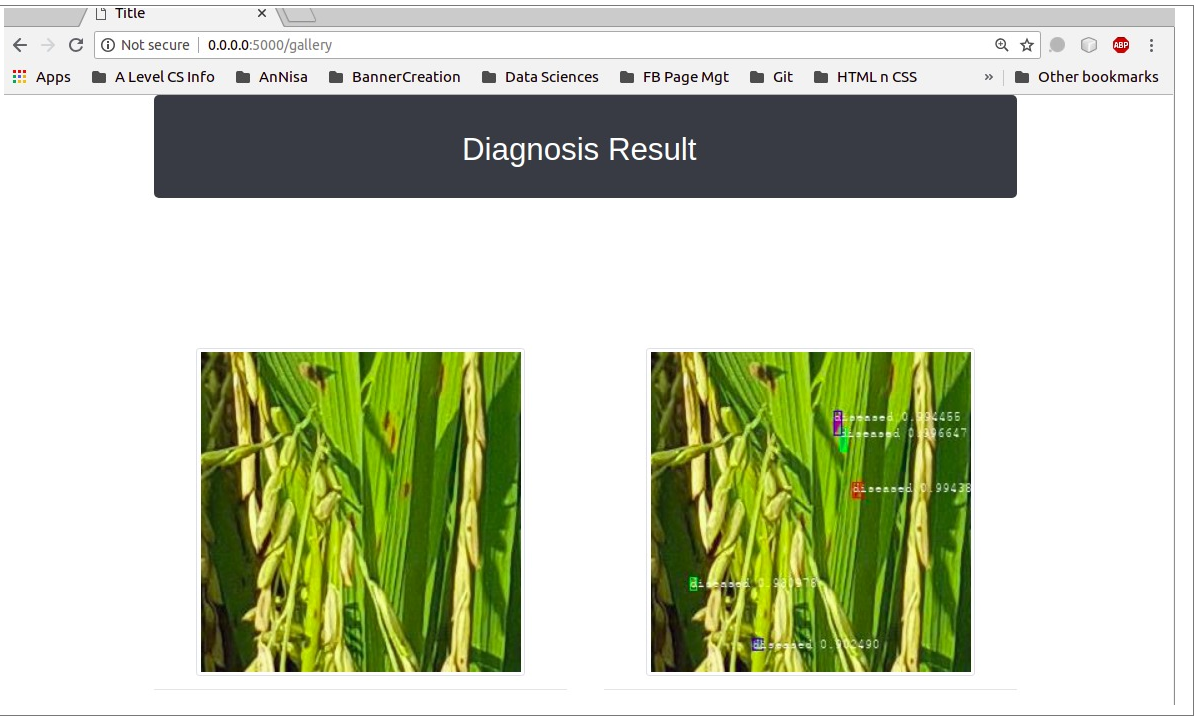} \\
  		(a) & (b)\\
    \end{tabular}  		
	\end{center}
	\caption{(a) Landscape mode rice crop image (b) Web Application.}
	\label{img:landscape_webapp}
\end{figure}

Rice crop images have been collected from the fertile land near River Chenab which flows from the Himachal Pradesh, India and merges into Indus River in Punjab, Pakistan. These images contain healthy and diseased instances of crop collected by actual farmers. Data has been collected in two phases, first during August, and second during September. Dataset contains around $1700$ images ($4032\times 1960$) collected using a mobile phone camera in day time during good light conditions. A sample image is shown in Fig.~\ref{img:landscape_webapp}(a). 

We first crop out roads and sky from the images, and then the blurred portions of the images to focus on informative portions. The images are then annotated as either healthy or diseased. Each diseased image is then further annotated to mark infected portion on the plant. Images are annotated using a Python based tool named LabelMe \citep{LabelMe}. This tool is used to draw bounding boxes and polygons on the diseased parts of the images. Out of $1700$ images, $600$ images were processed using LabelMe to include localized disease annotations. We have also developed a web application that allows farmers to see the automatically generated analysis report on their submitted images (see Fig.~\ref{img:landscape_webapp}(b)).

\section{Proposed Methodology} 
\label{sec:proposed_method}
In this work, we proposed a base-line CNN architecture and a Mask R-CNN based localization and classification strategy. The baseline CNN is designed for comparative purposes whereas Mask R-CNN is employed to illustrate the benefit of localized classification in early detection.

\subsection{Mask R-CNN}

Mask R-CNN \citep{MaskRCNN2017} is one of the variants of Region based Convolutional Neural Network. It was built on top of Faster R-CNN \citep{FasterRCNN2015}. It generates three outputs for each Region of Interest (RoI)/object i) class label, ii) a bounding box offset and iii) the object mask. Similar to Faster R-CNN, Mask R-CNN has two stages. It predicts potential regions for bounding boxes and extracts features from candidate regions. Along with bounding boxes and class labels, Mask R-CNN also outputs mask for each region and it has shown to outperform state-of-art-approaches on MS COCO dataset~\citep{COCO2014}. 

Our proposed Mask R-CNN contains the same ResNet 101~\citep{He2015} backbone as used in the vanilla implementation~\citep{MaskRCNN2017}. However, instead of $80$ regression heads (one for each category of MS COCO dataset), we  used two regression heads; one for healthy and the other for diseased. Proposed regions are predicted by passing image vector through convolutional and ROI layers, which then divide into parallel pipeline. One passes through Fully Connected (FC) layer and predicts class label and bounding box co-ordinates. Second pipeline uses convolution layer to output segmentation masks for each proposed region  (see Fig.~\ref{img:cnn_mask_rcnn_archi}(a)). 

Model is trained on $80\%$ of total images and is tested on remaining $20\%$ data. It is trained using $60$ epoch with a batch size of $2$. In every iteration, each image is passed through the network. The model is trained using gradient descent with momentum (learning rate $0.001$, learning momentum $0.9$). The model is trained with $1024\times 1024\times3$ size images with mini mask of size $56\times 56$.

\subsection{CNN}
\par
\begin{figure}[htbp]
	\begin{center}
	  \begin{tabular}{cc}
  		\includegraphics[width=0.5\columnwidth]{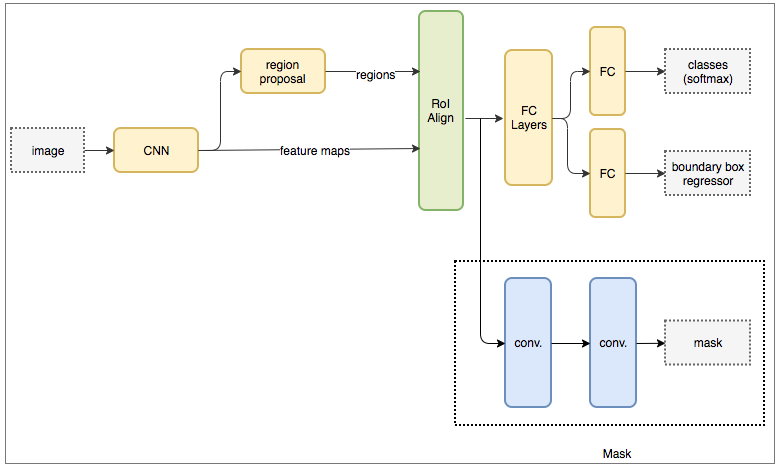}&
  		\includegraphics[width=0.5\columnwidth]{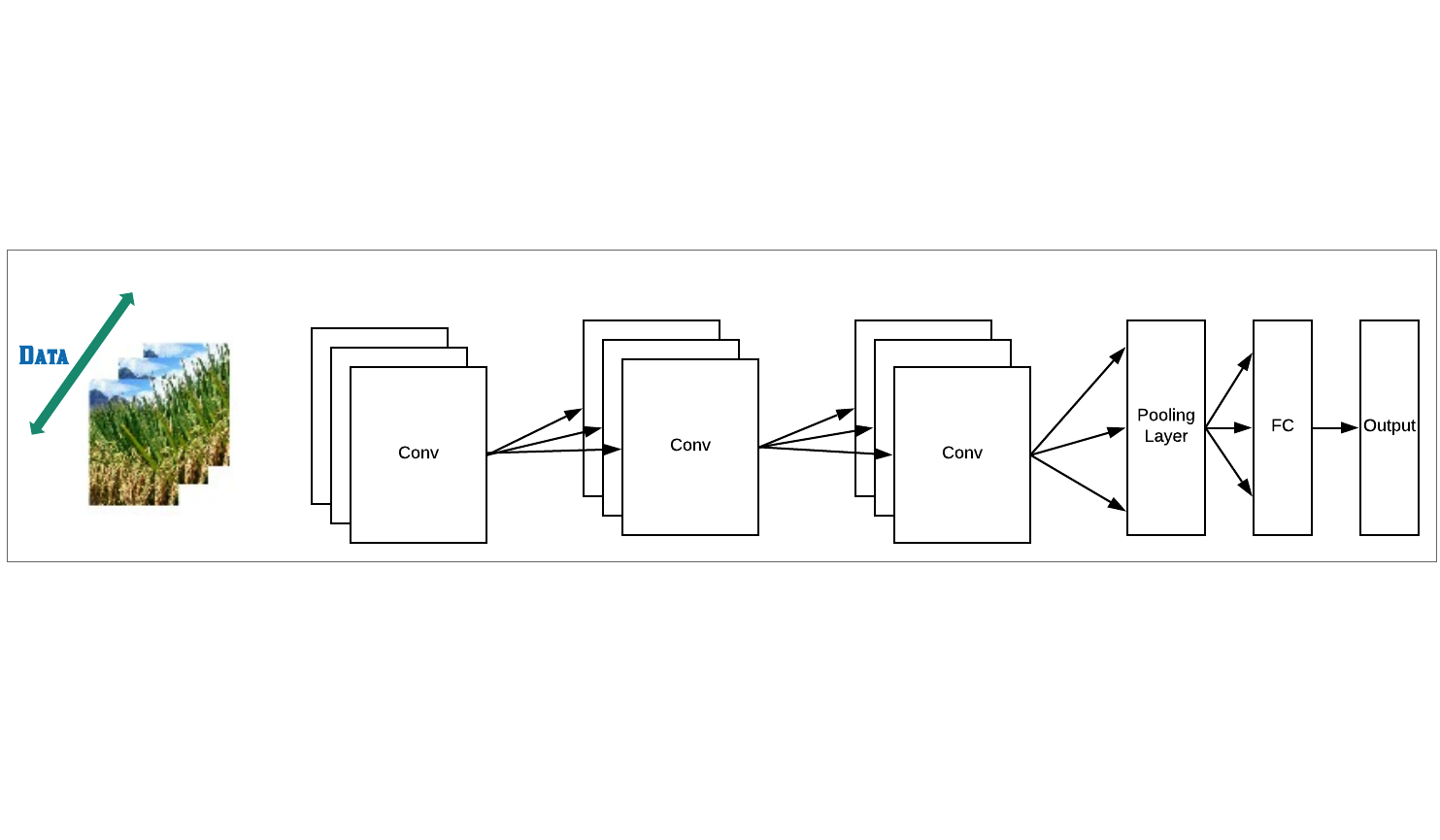} \\
  		(a) & (b)
  	  \end{tabular}	
	\end{center}
	\caption{(a) Mask R-CNN Architecture ~\citep{MaskRCNN2017}  (b) Baseline CNN Architecture}
	\label{img:cnn_mask_rcnn_archi}
\end{figure}

 
We also trained a baseline CNN-based classification architecture. The purpose of convolutional layer is to capture high level features of the data. Our CNN contains three convolutional layers with batch normalization in between activation layer, followed by pooling layers. Input to the fully connected layer is flatten out and class label is obtained from output layer. ReLU~\citep{ReLU} is used as activation function of hidden layers, whereas sigmoid is applied on the output layer (see Fig.~\ref{img:cnn_mask_rcnn_archi}(b)).
 
CNN model performed best when trained for $30$ epochs using Batch Gradient Descent with Adam as an optimizer.


\section{Results}\label{sec:results}
The developed models were tested on unseen dataset comprising of $340$ images containing equal number of healthy and diseased images. The performance comparison of the CNN and Mask R-CNN models was made using four standard performance measures. These include accuracy, precision, recall and F1-score. The measures were calculated using the formulae given in Table~\ref{tab:rslt}. Mask R-CNN shows $49.4\%$ improvement in accuracy, $36.29\%$ increase in precision, $45.3\%$ rise in recall and $41\%$ change in F1-scores as shown in Table~\ref{tab:rslt}. Furthermore, mean Average Precision for Mask R-CNN was $0.65$.

\begin{table}[htbp]
\caption{Comparison between CNN and Mask R-CNN architectures. (Keys: TP = True Positives, FP = False Positives, FN = False Negatives, TN = True Negatives)}
\begin{center}
\begin{tabular}{c|cccc}
\hline
& Accuracy & Precision & Recall & F1 score \\
& $\frac{TP+TN}{TP+TN+FP+FN}$ & $=\frac{TP}{TP+FP}$ & $\frac{TP}{TP+FN}$ & $\frac{2*Precision*Recall}{Precision + Recall}$ \\
\hline
\textbf{CNN}        & 0.584 & 0.598 & 0.662 & 0.628 \\
\textbf{Mask R-CNN} & 0.876 & 0.821 & 0.962 & 0.886 \\
\hline
\end{tabular}
\label{tab:rslt}
\end{center}
\end{table}

\begin{figure}[H]
\begin{center}
\begin{tabular}{cc}
\addvbuffer[-1pt]{\includegraphics[height=6cm,width=0.45\columnwidth]{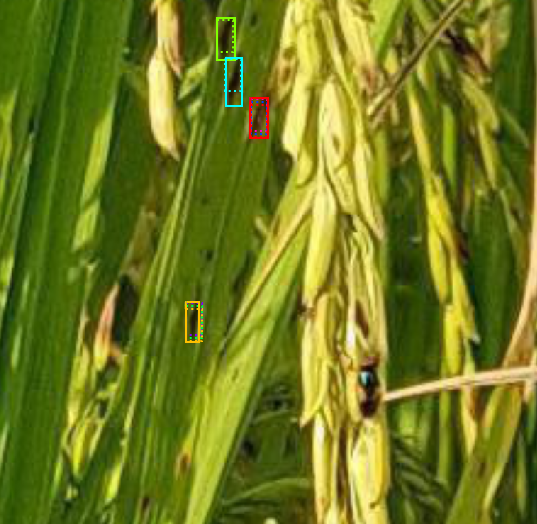}} &
\shortstack{%
\addvbuffer[-1pt]{\includegraphics[height=3cm,width=0.43\columnwidth]{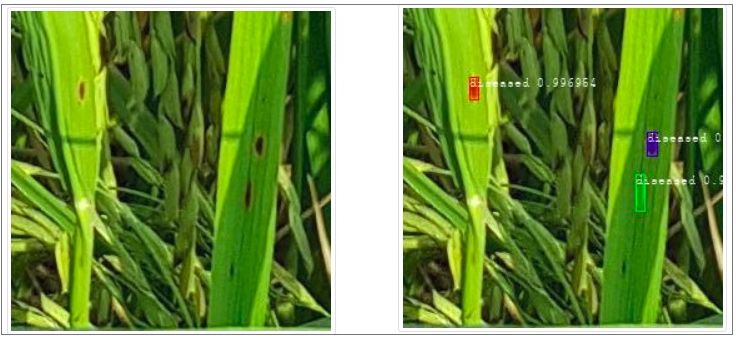}}\\
\addvbuffer[-1pt]{\includegraphics[height=3cm,width=0.45\columnwidth]{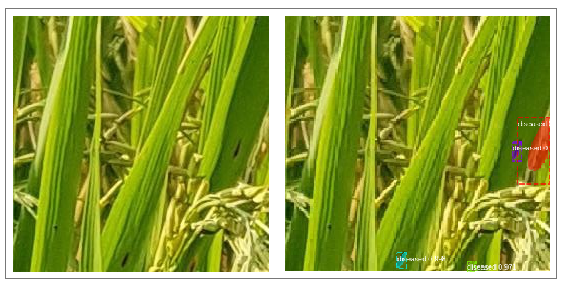}}
}\\
(a) & (b) \\
\end{tabular}
\end{center}
\caption{(a) Ground Truth (solid line) vs. Predicted Box (dashed line). (b) Input image and its predicted mask showing several correct localization along with a single misclassification.}
\label{img:iou_box_mask_pred_mis_class}
\end{figure}

Figure~\ref{img:iou_box_mask_pred_mis_class} shows sample result on an image taken from the dataset of rice plant images. Figure~\ref{img:iou_box_mask_pred_mis_class}(a), shows the comparison of the actual ground truth bounding box (solid line) with the predicted bounding box (dotted line). Instance based segmentation predicted by Mask-R-CNN with the respective probability is shown in Fig.~\ref{img:iou_box_mask_pred_mis_class}(b). It shows several correct localizations along with a single misclassification.

\section{Conclusion} \label{sec:conclusion}

Early disease prediction is important to help farmers to reduce the loss of crop yield especially for crops that are of high value. This can be accomplished by developing models that can easily be used by the farmers on their fields without requiring extensive technological know how. The present work outlines one such application for rice growers that takes images of the farm and processes it through Deep Learning (DL) model to predict the disease affected regions of the crop. The DL model is based on Mask R-CNN which identifies the localized disease portion of the plant along with its classification with accuracy of $87.6\%$. The paper shows that Mask R-CNN outperforms the standard CNN model in terms of accuracy, precision, recall, and F1-score. The model was trained on a locally collected dataset comprising of $1700$ images of both diseased and healthy plants. The data was collected using a simple mobile phone and was annotated manually. 

At present we are working on developing an easy to use multi-lingual mobile application that could be used on field by the farmers. We are also in the process of extending the dataset and the developed model into a multi-class disease detector and classifier.  
\newpage
\section*{Acknowledgment}
We thank National Agricultural Robotics Lab (NARL) working under National Centre of Robotics and Automation (NCRA) established at EME NUST, Islamabad with the financial assistance from Higher Education Commission of Pakistan.

\bibliography{ICLR2020}

\begin{thebibliography}{17}
\providecommand{\natexlab}[1]{#1}
\providecommand{\url}[1]{\texttt{#1}}
\expandafter\ifx\csname urlstyle\endcsname\relax
  \providecommand{\doi}[1]{doi: #1}\else
  \providecommand{\doi}{doi: \begingroup \urlstyle{rm}\Url}\fi

\bibitem[Fin(2018)]{FinanceSurvey2018}
Ministry of finance govt. of pakistan.
\newblock \url{http://www.finance.gov.pk/survey\_1718.html}, 2018.
\newblock Accessed: 2019-07-01.

\bibitem[PBS(2019)]{PBS2019}
Agriculture {S}tatistics, {P}akistan {B}ureau of {S}tatistics.
\newblock \url{http://www.pbs.gov.pk/content/agriculture-statistics}, 2019.
\newblock Accessed: 2019-05-28.

\bibitem[Barbedo(2016)]{b1}
Jayme Garcia~Arnal Barbedo.
\newblock A review on the main challenges in automatic plant disease
  identification based on visible range images.
\newblock \emph{Biosystems Engineering}, 144:\penalty0 52--60, 2016.

\bibitem[DeChant et~al.(2017)DeChant, Wiesner-Hanks, Chen, Stewart, Yosinski,
  Gore, Nelson, and Lipson]{Dechant2017}
Chad DeChant, Tyr Wiesner-Hanks, Siyuan Chen, Ethan Stewart, Jason Yosinski,
  Michael Gore, Rebecca Nelson, and Hod Lipson.
\newblock Automated identification of northern leaf blight-infected maize
  plants from field imagery using deep learning.
\newblock \emph{Phytopathology}, 107-11:\penalty0 1426--1432, 2017.

\bibitem[Fuentes et~al.(2017)Fuentes, Yoon, Kim, and Park]{A2017}
Alvaro Fuentes, Sook Yoon, Sang Kim, and Dong Park.
\newblock A robust deep-learning-based detector for real-time tomato plant
  diseases and pests recognition.
\newblock \emph{Sensors}, 17:\penalty0 2022, 09 2017.
\newblock \doi{10.3390/s17092022}.

\bibitem[Hallau et~al.(2017)Hallau, Neumann, Klatt, Kleinhenz, Klein, Kuhn,
  Röhrig, Bauckhage, Kersting, Mahlein, Steiner, and Oerke]{Hallu2017}
L.~Hallau, Marion Neumann, B.~Klatt, Benno Kleinhenz, T.~Klein, C.~Kuhn,
  Manfred Röhrig, Christian Bauckhage, K.~Kersting, Anne-Katrin Mahlein,
  U.~Steiner, and E.-C Oerke.
\newblock Automated identification of sugar beet diseases using smartphones.
\newblock \emph{Plant Pathology}, 06 2017.

\bibitem[He et~al.(2015)He, Zhang, Ren, and Sun]{He2015}
Kaiming He, Xiangyu Zhang, Shaoqing Ren, and Jian Sun.
\newblock Deep residual learning for image recognition.
\newblock \emph{arXiv preprint arXiv:1512.03385}, 2015.

\bibitem[He et~al.(2017)He, Gkioxari, Dollár, and Girshick]{MaskRCNN2017}
Kaiming He, Georgia Gkioxari, Piotr Dollár, and Ross Girshick.
\newblock Mask r-cnn.
\newblock 03 2017.

\bibitem[Joshi \& Jadhav(2015)Joshi and Jadhav]{Joshi2016}
Amrita Joshi and B.D. Jadhav.
\newblock Monitoring and controlling rice diseases using image processing
  techniques.
\newblock pp.\  471--476, 2015.

\bibitem[Kamilaris \& Prenafeta-Boldu(2018)Kamilaris and
  Prenafeta-Boldu]{DL2018}
A.~Kamilaris and F.~X. Prenafeta-Boldu.
\newblock Deep learning in agriculture: A survey.
\newblock \emph{Computers and Electronics in Agriculture}, 147:\penalty0 70 --
  90, 2018.

\bibitem[Lin et~al.(2014)Lin, Maire, Belongie, Bourdev, Girshick, Hays, Perona,
  Ramanan, Dollár, and Zitnick]{COCO2014}
Tsung-Yi Lin, Michael Maire, Serge~J. Belongie, Lubomir~D. Bourdev, Ross~B.
  Girshick, James Hays, Pietro Perona, Deva Ramanan, Piotr Dollár, and
  C.~Lawrence Zitnick.
\newblock Microsoft coco: Common objects in context.
\newblock 2014.

\bibitem[MIT \& Laboratory(2019)MIT and Laboratory]{LabelMe}
Computer~Science MIT and Artificial~Intelligence Laboratory.
\newblock A quick guide, labelme.
\newblock \url{http://labelme2.csail.mit.edu/Release3.0/index.php?message=1},
  2019.
\newblock Accessed: 2019-07-01.

\bibitem[Nair \& Hinton(2010)Nair and Hinton]{ReLU}
Vinod Nair and Geoffrey~E. Hinton.
\newblock Rectified linear units improve restricted boltzmann machines.
\newblock pp.\  807--814, 2010.

\bibitem[Phadikar et~al.(2013)Phadikar, Sil, and Das]{Phadikar2013}
Santanu Phadikar, Jaya Sil, and Asit~Kumar Das.
\newblock Rice diseases classification using feature selection and rule
  generation techniques.
\newblock \emph{Computer Electronics and Agriculture}, 90:\penalty0 76--85,
  2013.

\bibitem[Ren et~al.(2015)Ren, He, Girshick, and Sun]{FasterRCNN2015}
Shaoqing Ren, Kaiming He, Ross Girshick, and Jian Sun.
\newblock Faster r-cnn: Towards real-time object detection with region proposal
  networks.
\newblock \emph{IEEE Transactions on Pattern Analysis and Machine
  Intelligence}, 39, 06 2015.
\newblock \doi{10.1109/TPAMI.2016.2577031}.

\bibitem[Rumpf et~al.(2010)Rumpf, Mahlein, Steiner, Oerke, Dehne, and
  Plümer]{Rumpf2010}
T.~Rumpf, Anne-Katrin Mahlein, U.~Steiner, E.-C Oerke, H.-W Dehne, and Lutz
  Plümer.
\newblock Early detection and classification of plant diseases with support
  vector machines based on hyperspectral reflectance.
\newblock \emph{Computers and Electronics in Agriculture}, 74:\penalty0 91--99,
  2010.

\bibitem[Zhongzhi(2019)]{Zhongzhi2019}
Han Zhongzhi.
\newblock \emph{Using Deep Learning for Image-Based Plant Disease Detection},
  pp.\  295--305.
\newblock 09 2019.
\newblock ISBN 9780429289460.
\newblock \doi{10.1201/9780429289460-24}.

\end{thebibliography}
\bibliographystyle{ICLR2020}

\end{document}